\documentclass{article}

\usepackage[preprint]{neurips_2024}

\usepackage[utf8]{inputenc} 
\usepackage[T1]{fontenc}    
\usepackage{hyperref}       
\usepackage{url}            
\usepackage{booktabs}       
\usepackage{amsfonts}       
\usepackage{nicefrac}       
\usepackage{microtype}      
\usepackage{xcolor}         

\usepackage{amsmath}
\usepackage{amssymb}
\usepackage{mathtools}
\usepackage{amsthm}
\usepackage{algorithm}
\usepackage{algorithmicx}
\usepackage{algpseudocode}

\usepackage{cleveref}

\usepackage{cancel}

\usepackage{subfigure}

\usepackage{tcolorbox}
\tcbuselibrary{skins, breakable, theorems}

\usepackage{pifont}

\title{MoColl: Agent-Based Specific and General Model Collaboration for Image Captioning}
\author{%
  Pu Yang\thanks{First author.}\\
  the School of Mathematical Sciences\\
  Peking University\\
  \texttt{yang\_pu@pku.edu.cn} \\
  \AND
  Bin Dong\thanks{Corresponding author.} \\
  Beijing International Center for Mathematical Research \\
  Center for Machine Learning Research \\
  New Cornerstone Science Laboratory \\
  Peking University \\
  \texttt{dongbin@math.pku.edu.cn} \\
}

\begin{document}

\maketitle

\begin{abstract}
Image captioning is a critical task at the intersection of computer vision and natural language processing, with wide-ranging applications across various domains. 
For complex tasks such as diagnostic report generation, deep learning models require not only domain-specific image-caption datasets but also the incorporation of relevant general knowledge to provide contextual accuracy. 
Existing approaches exhibit inherent limitations: specialized models excel in capturing domain-specific details but lack generalization, while vision-language models (VLMs) built on large language models (LLMs) leverage general knowledge but struggle with domain-specific adaptation. 
To address these limitations, this paper proposes a novel agent-enhanced model collaboration framework, which we call \textbf{MoColl}, designed to effectively integrate domain-specific and general knowledge. 
Specifically, our approach is to decompose complex image captioning tasks into a series of interconnected question-answer subtasks. A trainable visual question answering (VQA) model is employed as a specialized tool to focus on domain-specific visual analysis, answering task-specific questions based on image content. Concurrently, an LLM-based agent with general knowledge formulates these questions and synthesizes the resulting question-answer pairs into coherent captions. Beyond its role in leveraging the VQA model, the agent further guides its training to enhance its domain-specific capabilities. 
Experimental results on radiology report generation validate the effectiveness of the proposed framework, demonstrating significant improvements in the quality of generated reports. 
\end{abstract}

\begin{figure}[htp]
    \centering
    \begin{minipage}[m]{0.48\textwidth}
    \centering
    \vspace{-6pt}
    \subfigure[Specialized encoder-decoder model, which maps from images to captions end-to-end. ]{
        \includegraphics[scale=0.55]{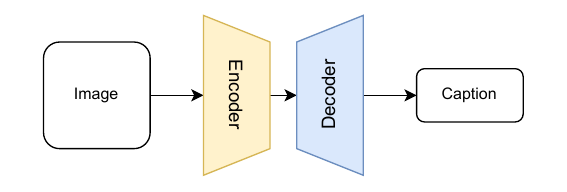}
        \label{subfig:specialized_model}
    }
    \subfigure[General vision language model, which takes the image and a task-dependent instruction as input and outputs the caption. ]{
        \includegraphics[scale=0.55]{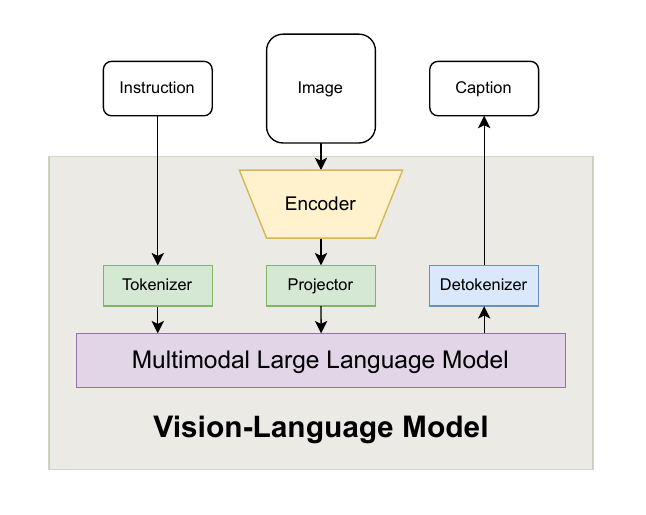}
        \label{subfig:general_model}
    }
    \end{minipage}
    \hspace{6pt}
    \begin{minipage}[m]{0.48\textwidth}
    \centering
    \subfigure[Our agent-enhanced model collaboration (MoColl) framework, where an LLM-based agent iteratively asks questions to a visual-question-answer model and gets answers from it. The agent provides a caption after enough question-answer pairs have been collected that are sufficient to understand the image. ]{
        \includegraphics[scale=0.55]{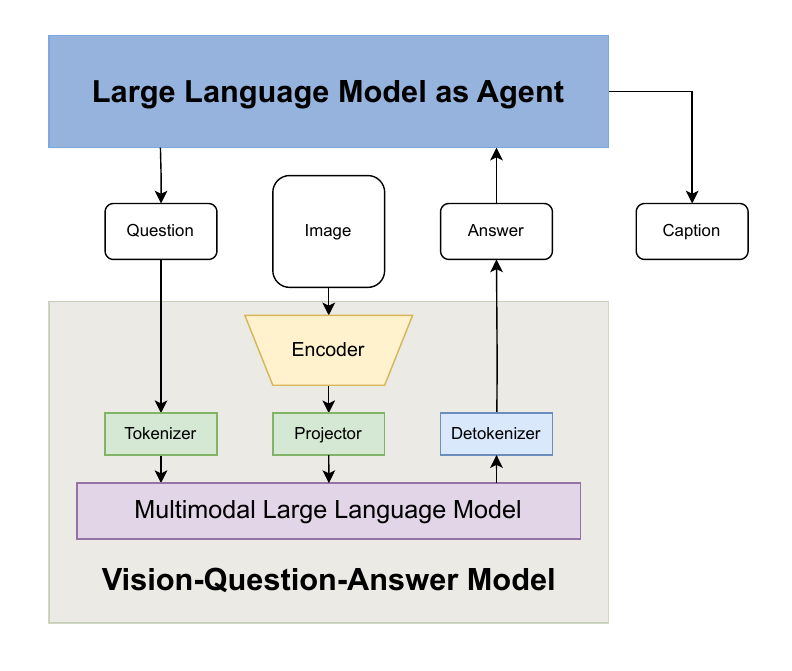}
        \label{subfig:tool_usage}
    }
    \end{minipage}
    \caption{Illustrative figures of three frameworks for image captioning. }
    \label{fig:intro}
\end{figure}

\section{Introduction}
\label{sec:intro}
Image captioning is a critical task in the intersection of computer vision and natural language processing, aiming to generate descriptive textual captions for given images. This task has profound implications for various applications such as radiology report generation \citep{monshi2020deep}.
Traditionally, approaches to image captioning have relied on specialized encoder-decoder models designed to learn the mapping from images to captions \citep{pmlr-v37-xuc15, Karpathy_2015_CVPR, Donahue_2015_CVPR, Anderson_2018_CVPR} (see \cref{subfig:specialized_model} for an illustration). These specialized models are typically direct and cost-effective, requiring relatively small amounts of data to function effectively. However, their primary drawback lies in their limited generalization capabilities, as they struggle to incorporate broader domain knowledge and often fail to adapt to diverse contexts. 

On the other hand, general models utilizing Vision-Language Models (VLMs) \citep{openai2023gpt4v, NEURIPS2023_6dcf277e, wang2023cogvlm, bai2023qwenvl, lu2024deepseek} present a compelling alternative (see \cref{subfig:general_model} for an illustration). VLMs, built upon the foundation of Large Language Models (LLMs), are extensively versatile, highly scalable, and imbued with substantial background knowledge. Therefore, these models have the ability to generate informative captions across a wide range of domains following instructions. However, general models face significant challenges when applied to specific-domain tasks. Two common strategies are used to adapt general models: In-Context Learning (ICL) and fine-tuning. ICL improves domain knowledge but does not enhance the model's ability to perform domain-specific tasks. On the other hand, fine-tuning can lead to catastrophic forgetting, where domain-specific training may degrade the model's ability to retain general knowledge \citep{mccloskey1989catastrophic, luo2023empirical}. As a result, neither approach effectively combines general knowledge with domain expertise for complex captioning tasks.

In this work, to address their limitations, we propose an agent-enhance model collaboration (MoColl) framework (see \cref{subfig:tool_usage} for an illustration) to integrate the domain-specific precision of specialized models with the contextual breadth of general models in complex image captioning tasks. To achieve this integration, we decompose the image captioning task into a series of question-answer subtasks, where a specialized visual question answering (VQA) model focuses on domain-specific visual analysis, and a general LLM-based agent formulates meaningful questions and synthesizes these question-answer pairs into coherent captions. 

Specifically, the VQA model serves as a domain-specific tool, designed to analyze visual content by answering targeted questions about the input images. This model can be trained on specific datasets to handle detailed and domain-relevant visual tasks efficiently. In parallel, the LLM-based agent, endowed with extensive general knowledge, plays a central role in planning and tool usage. It formulates contextually appropriate questions to query the VQA model and synthesizes the resulting question-answer pairs into coherent, domain-specific captions. Beyond this interplay, the agent actively enhances the VQA model's domain-specific capabilities. By generating synthetic question-answer pairs and filtering them based on ground truth captions, the agent curates high-quality training data to further refine the VQA model. 

Experimental results on radiology report generation demonstrate the effectiveness of this framework, showing significant improvements in the quality and specificity of generated captions. These findings highlight the potential of agent-guided systems in bridging the gap between domain expertise and general adaptability for complex, domain-specific tasks.

The main contributions of this paper are:
\begin{itemize}
    \item \textbf{Agent-enhanced model collaboration framework}: We propose a novel approach to image captioning by introducing a collaborative framework that combines an LLM-based agent and a specialized VQA model. 
    This design enables effective integration of domain-specific visual analysis with general knowledge of LLMs.
    \item \textbf{Agent-guided tuning algorithm}: Beyond integrating general LLM-based agent and specialized VQA model, we present an innovative idea where the agent actively contributes to the development of VQA model. 
    This adaptive tuning process significantly enhances the VQA model’s ability to address domain-specific visual tasks, improving its overall performance and flexibility in complex scenarios.
    \item \textbf{Experiments}: Our experiments demonstrate the state-of-the-art captioning performance on radiology report generation, validating the benefits of our proposed agent-enhanced model collaboration framework and agent-guided tuning procedure. 
\end{itemize}

The remainder of this paper is organized as follows. In \cref{sec:relat}, we discuss related works. In \cref{sec:method}, we introduce our framework of agent-guided tool tuning and usage. Experimental results are presented in \cref{sec:exp}, and finally, we conclude our findings in \cref{sec:conc}. 

\section{Related Work}
\label{sec:relat}
In this section, we review the key areas relevant to our proposed framework. In \cref{subsec:relat-special}, we discuss specialized models for image captioning, as they form the foundation for domain-specific tasks. In \cref{subsec:relat-vlm}, we discuss VLMs, as they integrate general knowledge into image-related tasks. They are both our main comparisons. In \cref{subsec:relat-llm}, we discuss the use of LLMs as agents, highlighting their ability to plan tasks, interact with tools, and act as optimizers, which are key functionalities in our framework. In \cref{subsec:relat-select}, we discuss approaches for selecting synthetic data, which are critical for the agent’s role in refining the VQA model by curating high-quality training data. 

We here only list works of direct relevance to ours, with an extensive reference list in \cref{appen-sec:relat}. 

\subsection{Specialized Model for Image Captioning}
\label{subsec:relat-special}
Specialized models for image captioning traditionally use an encoder-decoder pipeline \citep{pmlr-v37-xuc15, Karpathy_2015_CVPR, Donahue_2015_CVPR, Anderson_2018_CVPR}, a framework that has been extensively employed in various applications, e.g., radiology report generation (see \cref{appen-subsec:relat-report}). The encoder-decoder architecture is designed to learn the mapping from images to textual descriptions by encoding the visual information into a fixed-length vector and then decoding this vector into a coherent sequence of words. 

However, these models may struggle with understanding and incorporating broader contextual knowledge that is not explicitly present in the training data. As a result, their generalization capabilities are limited, making it difficult for them to adapt to diverse types of medical images beyond the specific domain they were trained on. 

\subsection{Vision-Language Model}
\label{subsec:relat-vlm}
Vision-Language Models (VLMs) such as GPT-4V \citep{openai2023gpt4v}, LLaVA \citep{NEURIPS2023_6dcf277e, Liu_2024_CVPR}, DeepSeek-VL \citep{lu2024deepseek}, CogVLM \citep{wang2023cogvlm}, and Qwen-VL \citep{bai2023qwenvl}, are usually based on LLMs such as GPT-4 \citep{achiam2023gpt}, Llama-2 \citep{touvron2023llama}, DeepSeek LLM \citep{bi2024deepseek}, Vicuna \citep{vicuna2023}, and Qwen-LM \citep{bai2023qwen} respectively. These models integrate visual and textual data, treating tokenized images as another form of language input. This allows them to leverage background knowledge, which in turn provides comprehensive understanding and generation capabilities. However, despite their extensive background knowledge, general VLMs often lack the granularity and specificity in domain expertise that specialized models possess, which manifests as a less detailed grasp of specialized knowledge.

To elevate the capabilities of general models in domain-specific tasks, our framework requires a tool that not only excels in specific-domain proficiency but also interacts effectively with general LLM-based agent. We address this by employing a VQA \citep{Antol_2015_ICCV} model, which we fine-tune on domain-specific datasets to tailor its capabilities. Meanwhile, VQA models, which accept images and natural language questions as inputs and generate corresponding answers, offer the flexibility to adeptly manage image-based queries posed by the agent. Although the VQA model is a specialized form of a VLM, within our framework, we utilize it as a specialized tool. 

\subsection{Large Language Model as Agent}
\label{subsec:relat-llm}
Recent research, such as MetaGPT \citep{hong2024metagpt} and AutoGPT \citep{AutoGPT}, has found that LLMs can be employed as agents, acting as core controllers for executing complex tasks. In particular, their application extends across multiple domains of functionality, each illustrating the versatility of their capabilities as autonomous systems.

\paragraph{Planning.} LLM-based agents excel in task planning by decomposing tasks into smaller, sequential steps. Techniques such as Chain-of-Thought (CoT) \citep{NEURIPS2022_9d560961}, Tree-of-Thought \citep{NEURIPS2023_271db992}, and LLM+P \citep{liu2023llm+} enable these agents to plan and execute complex workflows. 

\paragraph{Tool usage.} A crucial aspect of LLM-based agents is their ability to utilize external tools effectively. Systems like MRKL \citep{karpas2022mrkl}, TALM \citep{parisi2022talm}, Toolformer \citep{NEURIPS2023_d842425e}, and ChatGPT Plugins \citep{GPTPlugin} enable LLMs to interact with various tools to gather information, perform computations, and generate solutions. This capability significantly enhances their versatility and effectiveness in problem-solving.

\paragraph{LLM as optimizer.} One of the most significant advancements in utilizing LLMs as agents is their role as optimizers. LLMs excel in optimizing prompts \citep{yang2024large}, where they generate and refine input queries to maximize the effectiveness and relevance of their responses. What's more, LLMs have also demonstrated a remarkable ability in refining machine learning models by tuning hyperparameters \citep{zhang2023using, liu2024large}, suggesting architectural improvements \citep{zheng2023can}, improving loss functions \citep{song2023self, NEURIPS2023_034d7bfe}, and even guiding exploration in reinforcement learning \citep{pmlr-v202-du23f}. Building on this work, we propose to innovate further by leveraging LLMs to optimize \textit{data}, i.e., dynamically generating and refining training datasets, particularly through the creation of targeted visual-question-answer data that address specific weaknesses of VQA models. 

\subsection{Synthetic data with selection}
\label{subsec:relat-select}
Recent studies \citep{mitra2024agentinstruct} have demonstrated that high-quality synthetic data is crucial for the post-training enhancement of language models. Although training with synthetic data suffers from model collapse \citep{shumailov2023curse}, data selecting techniques utilize a verifier that can improve the quality of synthetic data. For instance, golden verifiers, such as the Python interpreter, are used for code generation \citep{haluptzok2022language}, and symbolic deduction engines are used for Olympiad geometry problems \citep{trinh2024solving}. When a golden verifier does not exist, some studies employ heuristic verifiers, such as training a verification model with high-quality data \citep{pmlr-v162-li22n}, designing critic prompts \citep{wang2022self, wei2023magicoder}, and leveraging a self-critique mechanism \citep{putta2024agent}. In this work, we prompt a strong foundation model to select the synthetic question-answer data. 

\section{Methods}
\label{sec:method}
In this section, we introduce our proposed agent-enhanced model collaboration framework and agent-guided tuning framework. 
We first outline the model collaboration framework in \cref{subsec:generate}, where the agent and the VQA model are combined to decompose the captioning task into subtasks of questions and answers. While this base setup allows for effective interaction between the agent and the VQA model, its performance is limited when the VQA model lacks domain-specific training on a task-specific VQA dataset. To address this, in \cref{subsec:train}, we introduce the concept of LLM as an optimizer and describe how this training process refines the VQA model, enabling it to perform better in domain-specific scenarios.

\subsection{Model Collaboration Framework}
\label{subsec:generate}
This framework decomposes the captioning process into manageable subtasks, where the agent iteratively formulates targeted questions for the VQA model, gathers answers, and synthesizes them into accurate and coherent captions. It is detailed as follows and illustrated in \cref{subfig:tool_usage}. 

Specifically, the caption generation process can be divided into questioning step and captioning step. In the questioning step, the agent iteratively asks a question to the VQA model with reference to previous conversations (if any), and then the VQA model will feedback an answer according to the given image, i.e.
\begin{equation} \label{eq:question}
    q_{ij} = \text{Agent}_{\text{question}} (\{(q_{ik}, a_{ik})\}_{k<j}) , \quad j = 1, 2, \cdots, 
\end{equation}
\begin{equation} \label{eq:answer}
    a_{ij} = \text{VQA} (q_{ij}, x_i), \quad j = 1, 2, \cdots, 
\end{equation}
where $q_{ij}$ is a question, $a_{ij}$ is an answer, and $x_i$ is the image(s). The captioning step is after enough question-answer pairs have been collected or a stopping criterion is achieved, where we prompt the agent to provide the caption according to the sequence of questions and answers, i.e.
\begin{equation*} 
    \hat{c}_i = \text{Agent}_{\text{caption}} (\{(q_{ij}, a_{ij})\}_j).
\end{equation*}
Thus, we get the generated caption $\hat{c}_i$. 

We summarize the inference process in \cref{alg:inference}. 

\begin{algorithm}[t]
\caption{Model Collaboration (Inference)}
\label{alg:inference}
\begin{algorithmic}[1]
\Require the image $x$, the agent $\text{Agent}_\text{question}$ and $\text{Agent}_\text{caption}$, the VQA model, the maximum number of questions $M$
\For{$j \gets 1 \ \text{to} \ M$}
\State $q_j \gets \text{Agent}_{\text{question}}(\{(q_k, a_k)\}_{k<j})$ \Comment{question}
\If{$q_j$ is \textbf{None}}
\State \textbf{break} \Comment{enough information}
\EndIf
\State $a_j \gets \text{VQA} (x, q_j)$ \Comment{answer}
\EndFor
\State $\hat{c} \gets \text{Agent}_{\text{caption}}(\{(q_k, a_k)\}_{k<j})$ \Comment{caption}
\State \Return{$\{ \{(q_k, a_k)\}_{k<j}, \hat{c}\}$}
\end{algorithmic}
\end{algorithm}

In-context learning (ICL) \citep{NEURIPS2020_1457c0d6} serves as one of the most effective techniques in prompt engineering to enhance the capacity of LLM. We can provide some few-shot examples of captions to give the agent a more explicit perception of the scene it is in. One of the ICL strategies is to randomly select captions in the training set as examples. Another more advanced example choice strategy, inspired by Retrieval Augmented Generation (RAG) \citep{NEURIPS2020_6b493230}, is to finds the captions corresponding to images most similar to the query image as examples. In particular, we measure the similarity between images with the cosine similarity metric where images are embedded with the pretrained CLIP Vision Encoder \citep{pmlr-v139-radford21a}. 

However, without domain-specific training on a task-specific VQA dataset, the VQA model lacks the necessary domain knowledge to handle specialized tasks effectively. This limitation is evident in our experimental results (see \cref{subsec:exp-main}), where the untrained VQA model underperforms in domain-specific scenarios. To address this, the following subsection introduces our training process, where the agent not only utilizes the VQA model but also enhances it by guiding its domain-specific training through synthetic data generation and selection.

\begin{figure}[t]
    \centering
    \subfigure[The whole training procedure. ]{
        \centering
        \includegraphics[width=\linewidth]{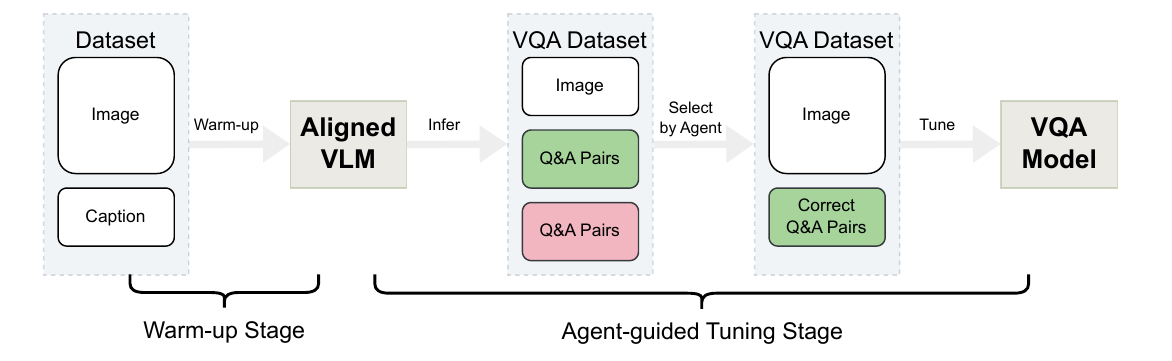}
        \label{subfig:train}
    }
    \subfigure[Warm-up stage. This stage aligns the image features to their textual word embedding in the pre-trained LM. The encoder and LM are frozen and only the projector can be updated. We train it via auto-regression, i.e., predicting the ground truth caption through a task-dependent instruction and the image(s). ]{
        \centering
        \includegraphics[scale=0.48]{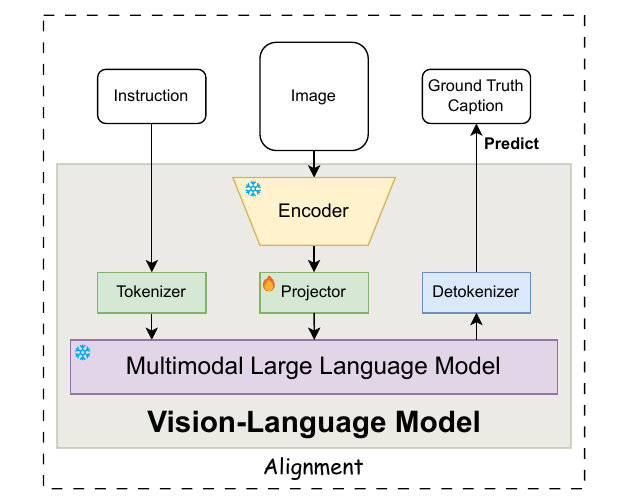}
        \label{subfig:warm-up}
            }
    \hspace{10pt}
    \subfigure[Agent-guided tuning stage. This stage consists of three steps. In the `Inference' step, the LLM-based agent generates a large amount of question-answer pairs by invoking the VQA model. In the `Selection' step, the agent will select high-quality synthetic data and filter out low-quality synthetic data according to corresponding ground truth captions. In the `Training' step, the VQA model is trained on these selected data via auto-regression. ]{
        \centering
        \includegraphics[scale=0.45]{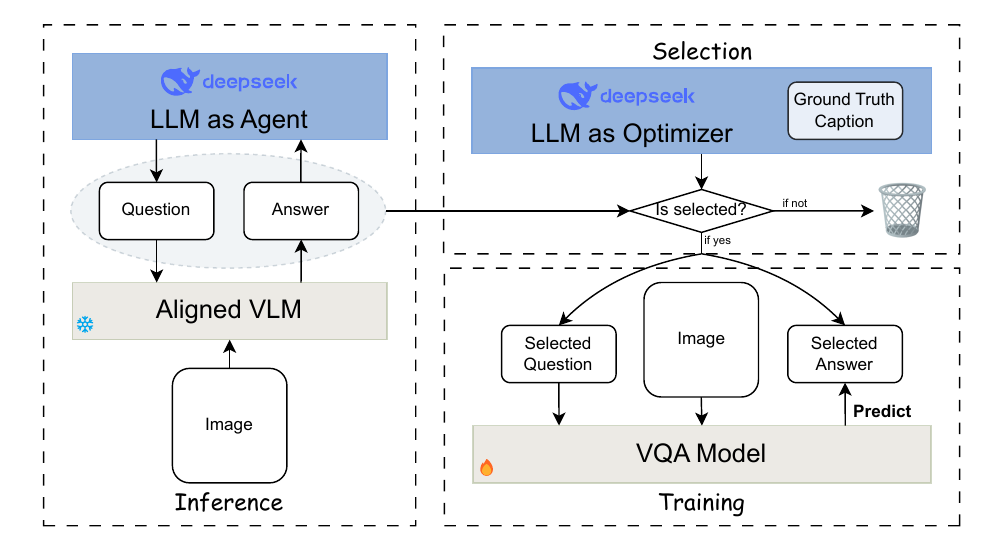}
        \label{subfig:tuning}
    }
    \caption{Illustration of the training procedure of our method. }
    \label{fig:train}
\end{figure}

\subsection{Training Procedure}
\label{subsec:train}
The training procedure consists of the following 2 stages. The first stage is the warm-up stage, where we align the image features to their textual word embeddings in the pre-trained LM. The second stage is the main training stage called agent-guided tuning, where we fine-tune the VQA model with visual-question-answer data generated by interaction between the agent and VQA model and selected by the agent. It is detailed as follows and the whole training procedure is illustrated in \cref{subfig:train}.

\subsubsection{Stage 1: Warm up} 
\label{subsec:warm-up}
Following the alignment process in LLaVA-Med \citep{NEURIPS2023_5abcdf8e}, we first align the image features to their textual word embeddings in the pre-trained LM. Specifically, we convert the image-caption pair data into the image-instruction-caption tuple, where the instruction simply presents the task of describing the image, e.g. `Provide a diagnostic report based on the given image(s).'. Here we use auto-regressive loss that asks the VQA model to predict the caption according to the given image and instruction, i.e., $p(c_i \mid x_i, \mathcal{I})$, where $c_i$ is its corresponding ground truth caption, and $\mathcal{I}$ is the task-dependent instruction. We keep both the visual encoder and LM weights frozen, and only update the projection network. This stage is illustrated in \cref{subfig:warm-up}. 

\subsubsection{Stage 2: Agent-Guided Tuning} 
\label{subsec:tuning}
This stage aims to fine-tune the VQA model through the guidance of the agent. The core idea is that the agent curates high-quality training data to refine the VQA model by generating synthetic question-answer pairs within the model collaboration framework and filtering them using ground truth captions. It consists of three steps: (1) inference, (2) selection, and (3) training. 

We summarize the agent-guided tuning process in \cref{alg:train} and illustrate it in \cref{subfig:tuning}. Details of these three steps are shown as follows. 

\paragraph{Inference step.} This step aims to synthesize many synthetic question-answer pairs from image-caption pairs in the training dataset. Specifically, we simulate the questioning step of the model collaboration framework (as detailed in \cref{subsec:generate}) in which the agent continuously asks questions to the VQA model and receives answers (see \cref{eq:question} and \cref{eq:answer}). Similarly, the iteration will not stop until the agent ensures that enough image information has been collected through these question-answer pairs or achieves a stopping criterion. The inference process is in lines 1-12 of \cref{alg:train}.

\paragraph{Selection step.} This step aims to select high-quality synthetic question-answer pairs and thus build a VQA dataset. Inspired by recent work on selecting synthetic data, we propose that the agent can be used to improve the performance of the VQA model via selected synthetic data. Specifically, we instruct the agent to select correct question-answer pairs according ground truth captions. Similar to the process of a teacher looking at a standard answer to correct a student's error, the selection agent can find out inappropriate questions from the above questioning agent and inaccurate answers from the VQA model, and then discards them. Subsequently, it remains high-quality question-answer pairs for the next training step, formalized as
\begin{equation*}
    \{(q_{ij}, a_{ij}) \mid \text{Agent}_{\text{select}} (q_{ij}, a_{ij}, c_i) = \text{True}\}. 
\end{equation*}
The selection process is in lines 13-18 of \cref{alg:train}. 

\paragraph{Training step.} This step aims to fine-tune the VQA model with the high-quality VQA dataset. It improves the performance of the VQA model and ensures that the tool remains effective and adaptable. Specifically, we use auto-regressive loss, i.e., $p(a_j \mid x_j, q_j; \text{VQA})$, where $(x_j, q_j, a_j)$ is the selected image-question-answer tuple. The visual encoder weights are still frozen, and the LM and the projection network can be updated. The training process is in lines 19-24 of \cref{alg:train}.

Overall, with the whole training procedure, we not only integrated domain-specific knowledge but also significantly enhanced the VQA model's capabilities. In the following experimental section, We will conduct a series of experiments to demonstrate the advantages of our methods.

\begin{algorithm}[t]
\caption{Agent-Guided Tuning (Stage 2 of Training)}
\label{alg:train}
\begin{algorithmic}[1]
\Require an image captioning dataset $\mathcal{D} = \{(x_i, c_i)\}_{i=1}^N$, an aligned VLM, the agent $\text{Agent}_\text{question}$ and $\text{Agent}_\text{select}$, the maximum number of questions $M$

\hspace{-20pt}// Step 1: Infer (see \cref{alg:inference}) without captioning.
\State Initialize Memories $\mathcal{M} \gets \text{Set}()$
\For{$i \gets 1 \ \text{to} \ N$}
\State Sample $(x_i, c_i) \sim \mathcal{D}$
\For{$j \gets 1 \ \text{to} \ M$}
\State $q_{ij} \gets \text{Agent}_{\text{question}} (\{(q_{ik}, a_{ik})\}_{k<j})$ \Comment{question}
\If{$q_{ij}$ is \textbf{None}}
\State \textbf{break}  \Comment{enough information}
\EndIf
\State $a_{ij} \gets \text{VLM} (q_{ij}, x_i)$ \Comment{answer}
\State Add $( x_i, q_{ij}, a_{ij}, c_i )$ into $\mathcal{M}$
\EndFor
\EndFor

\hspace{-20pt}// Step 2: Select according to the ground truth caption. 
\State Initialize the VQA dataset $\mathcal{D}^{\text{VQA}} \gets \text{Set}()$
\For{$(x_i, q_{ij}, a_{ij}, c_i) \ \text{in} \ \mathcal{M}$}
\If{$\text{Agent}_{\text{select}} (q_{ij}, a_{ij}, c_i)$ is \textbf{true}} 
\State Add $(x_i, q_{ij}, a_{ij})$ into $\mathcal{D}^{\text{VQA}}$ \Comment{select}
\EndIf
\EndFor

\hspace{-20pt}// Step 3: Train the VQA model. 
\State Initialize the VQA model $\text{VQA}_\theta \gets \text{VLM}$
\For{$i \gets 1 \ \text{to} \ \text{max\_steps}$}
\State Sample a mini-batch $\{(x_j, q_j, a_j)\}_{j=1}^B \sim \mathcal{D}^{\text{VQA}}$
\State $\mathcal{L}(\theta) = \frac{1}{B} \sum_{j=1}^{B} \log p(a_j \mid x_j, q_j; \text{VQA}_\theta)$ \Comment{auto-regressive loss}
\State Update $\theta$ through $\mathcal{L}(\theta)$ with an optimizer, e.g., ADMM 
\EndFor
\State \Return{$\text{VQA}_\theta$ model}
\end{algorithmic}
\end{algorithm}

\section{Experiments}
\label{sec:exp}

\subsection{Implementation}
\label{subsec:imp}

\paragraph{Datasets.} We verify our proposed methods on two widly used datasets: (1) IU-Xray \citep{demner2016preparing} \footnote{\url{https://openi.nlm.nih.gov/}}, which consists of 3955 reports and 7,470 chest X-ray images; and (2) MIMIC-CXR \citep{johnson2019mimic} \footnote{\url{https://physionet.org/content/mimic-cxr-jpg/2.1.0/}}, which consists of more than 206,000 reports and 473,000 chest X-ray images. Details of data preprocessing are provided in \cref{appen-subsec:data}. 

\paragraph{Setup.} Details of data preprocessing: (1) \textbf{Within-dataset evaluation}, where we assess the model on the test set of MIMIC-CXR, and (2) \textbf{Cross-dataset generalization}, where we test the model on the IU-Xray dataset to examine its ability to generalize from one training set to a similar but distinct test set.

\paragraph{Metrics.} We employ four commonly used natural language processing evaluation metrics to assess captioning performance: BLEU\{1-4\} \citep{papineni2002bleu}, Meteor \citep{denkowski2011meteor}, ROUGE-L \citep{lin2004rouge}, and Cider \citep{Vedantam_2015_CVPR}. 

\paragraph{Our methods.} We detail the specialized VQA tool and LLM-based agent employed in our framework. For the VQA model, we use LLaVA \citep{NEURIPS2023_6dcf277e}, a visual-language conversation model, as our VQA model. The architecture remains unchanged, consisting of a frozen image encoder, a projection network, and a pre-trained causal Language Model (LM). Specifically, we use CLIP Vision Encoder \citep{pmlr-v139-radford21a} as the image encoder, which outputs a vector representing the features of the input image. The image features are then mapped into visual tokens via a randomly initialized three-layer Multi-Layer Perceptron (MLP), which serves as an intermediary, projecting the image information into a "foreign language" understood by the LM. The pre-trained LM is Llama-3.1-8B \citep{dubey2024llama}. For the agent, we adopt DeepSeek-V2.5 \citep{liu2024deepseek}. Furthermore, we demonstrate our model collaboration (MoColl) framework with two tools: the aligned VLM and the tuned VQA model. The former is trained only through a warm-up stage, while the latter is trained from the aligned VLM through an agent-guided tuning stage. Additional implementation details can be found in \cref{appen-subsec:method}.

\paragraph{Baseline methods.} We evaluate our method against three categories of baseline approaches, as depicted in \cref{fig:intro}. (1) \textbf{Specialized encoder-decoder models}: R2Gen \citep{chen-emnlp-2020-r2gen}, Joint-TriNet \citep{yang2021joint}, R2GenCMN \citep{chen-etal-2021-cross-modal}, XProNet \citep{wang2022cross}, and M2KT \citep{yang2023radiology}. (2) \textbf{General vision language models}: LLaVA-Med \citep{NEURIPS2023_5abcdf8e}, and LLaVA-1.5-7b \citep{Liu_2024_CVPR}, DeepSeek-VL2 \citep{lu2024deepseek}, Qwen2-VL-72B-Instruct \citep{bai2023qwenvl} (denoted as Qwen2-VL), our aligned VLM, and supervised fine-tuned VLM. (3) \textbf{Model collaboration framework with VQA models}: LLaVA-Med and LLaVA-1.5-7b. Implementation details of each baseline method are provided in the \cref{appen-sec:comp}. 

\subsection{Main Results}
\label{subsec:exp-main}

\begin{table}[t]
    \centering
    \caption{Main results of within-dataset evaluation. The best results (highest value for each metric) among the compared algorithms are shown in bold numbers. The compared methods are grouped into specialized encoder-decoder models (top section), general VLMs (middle section), and our MoColl framework with various VQA models (bottom section). Our methods are MoColl framework with the algined VLM and the agent-guided tuned VQA model, which belong to the bottom section. }
    \resizebox{\linewidth}{!}{
    \begin{tabular}{l|ccccccc}
    \toprule
         & BLEU1 & BLEU2 & BLEU3 & BLEU4 & Meteor & ROUGE-L & Cider \\ \midrule
        Joint-TriNet \citep{yang2021joint} & 0.2615 & 0.1127 & 0.0558 & 0.0300 & 0.0923 & 0.2051 & 0.0404 \\
        R2Gen \citep{chen-emnlp-2020-r2gen} & 0.1892 & 0.1172 & 0.0784 & 0.0556 & 0.0910 & 0.2304 & 0.0735 \\
        R2GenCMN \citep{chen-etal-2021-cross-modal} & 0.2290 & 0.1361 & 0.0882 & 0.0609 & 0.1007 & 0.2269 & 0.0838 \\
        XproNet \citep{wang2022cross} & 0.1608 & 0.0990 & 0.0662 & 0.0470 & 0.0865 & 0.2289 & 0.0797 \\
        M2KT \citep{yang2023radiology} & 0.2735 & 0.1257 & 0.0659 & 0.0384 & 0.1117 & 0.1742 & 0.0439 \\ \midrule
        LLaVA-Med \citep{NEURIPS2023_5abcdf8e} & 0.1475 & 0.0553 & 0.0197 & 0.0071 & 0.0646 & 0.1046 & 0.0079 \\ 
        LLaVA-v1.5-7B \citep{Liu_2024_CVPR} & 0.1519 & 0.0645 & 0.0264 & 0.0106 & 0.0747 & 0.1124 & 0.0089 \\ 
        Qwen2-VL \citep{bai2023qwenvl} & 0.1298 & 0.0482 & 0.0230 & 0.0120 & 0.0542 & 0.0925 & 0.0191 \\
        Deepseek-VL2 \citep{lu2024deepseek} & 0.0959 & 0.0449 & 0.0226 & 0.0121 & 0.0811 & 0.1173 & 0.0189 \\
        Aligned VLM & 0.2400 & 0.1293 & 0.0747 & 0.0450 & 0.0938 & 0.1614 & 0.0480 \\ 
        Fine-tuned VLM & 0.1760 & 0.0952 & 0.0561 & 0.0353 & 0.0783 & 0.1589 & 0.0464 \\
        \midrule
        MoColl + LLaVA-Med & 0.1865 & 0.0768 & 0.0335 & 0.0154 & 0.0664 & 0.1128 & 0.0212 \\
        MoColl + LLaVA-v1.5-7B & 0.1890 & 0.0774 & 0.0333 & 0.0146 & 0.0659 & 0.1163 & 0.0261 \\
        MoColl + aligned VLM (Ours) & 0.2515 & 0.1290 & 0.0746 & 0.0456 & 0.0947 & 0.1906 & 0.0566 \\
        MoColl + tuned VQA model (Ours) & \bf{0.2749} & \bf{0.1454} & \bf{0.0897} & \bf{0.0655} & \bf{0.1126} & \bf{0.2349} & \bf{0.0932} \\
    \bottomrule
    \end{tabular}
    }
    \label{tab:sft-mimic-cxr}
\end{table}

\begin{table}[t]
    \centering
    \caption{Main results of cross-dataset generation. }
    \resizebox{\linewidth}{!}{%
    \begin{tabular}{l|ccccccc}
    \toprule
         & BLEU1 & BLEU2 & BLEU3 & BLEU4 & Meteor & ROUGE-L & Cider \\ \midrule
        Joint-TriNet \citep{yang2021joint} & 0.2313 & 0.0855 & 0.0420 & 0.0230 & 0.0860 & 0.2161 & 0.0345 \\
        R2Gen \citep{chen-emnlp-2020-r2gen} & 0.3480 & 0.2057 & 0.1240 & 0.0772 & 0.1210 & 0.2494 & 0.1048 \\
        R2GenCMN \citep{chen-etal-2021-cross-modal} & 0.3480 & 0.2120 & 0.1342 & 0.0879 & 0.1329 & 0.2675 & 0.1462 \\
        XproNet \citep{wang2022cross} & 0.2853 & 0.1678 & 0.1007 & 0.0631 & 0.1069 & 0.2304 & 0.1075 \\
        M2KT \citep{yang2023radiology} & 0.2260 & 0.0961 & 0.0471 & 0.0252 & 0.1253 & 0.1889 & 0.0152 \\ \midrule
        LLaVA-Med \citep{NEURIPS2023_5abcdf8e} & 0.1068 & 0.0425 & 0.0192 & 0.0094 & 0.0704 & 0.1017 & 0.0030 \\ 
        LLaVA-v1.5-7B \citep{Liu_2024_CVPR} & 0.1296 & 0.0598 & 0.0304 & 0.0152 & 0.0930 & 0.1329 & 0.0040 \\ 
        Qwen2-VL \citep{bai2023qwenvl} & 0.1439 & 0.0718 & 0.0394 & 0.0210 & 0.1172 & 0.1770 & 0.0333 \\
        Deepseek-VL2 \citep{lu2024deepseek} & 0.0689 & 0.0350 & 0.0194 & 0.0107 & 0.0775 & 0.1648 & 0.0292 \\
        Aligned VLM & 0.3487 & 0.2109 & 0.1333 & 0.0827 & 0.1345 & 0.2619 & 0.1326 \\ 
        Fine-tuned VLM & 0.2702 & 0.1585 & 0.1015 & 0.0635 & 0.1106 & 0.2306 & 0.1231 \\ \midrule
        MoColl + LLaVA-Med & 0.2636 & 0.1368 & 0.0784 & 0.0452 & 0.1041 & 0.1962 & 0.0743 \\
        MoColl + LLaVA-v1.5-7B & 0.2411 & 0.1222 & 0.0693 & 0.0390 & 0.1009 & 0.1923 & 0.0648 \\
        MoColl + aligned VLM (Ours) & 0.3262 & 0.2033 & 0.1339 & 0.0861 & 0.1361 & 0.2722 & 0.1560 \\
        MoColl + tuned VQA model (Ours) & \bf{0.3515} & \bf{0.2158} & \bf{0.1413} & \bf{0.0890} & \bf{0.1372} & \bf{0.2822} & \bf{0.1749} \\
    \bottomrule
    \end{tabular}
    }
    \label{tab:inter}
\end{table}

We show the \textbf{within-dataset evaluation} results in \cref{tab:sft-mimic-cxr} and the \textbf{cross-dataset generation} results in \cref{tab:inter}. Key findings from the analysis are: 
\begin{itemize}
    \item \textbf{Best Performance of Our Method}: Across both testing scenarios, the MoColl framework, particularly with the agent-guided tuning algorithm, consistently outperforms all baseline methods across all metrics. This underscores the efficacy of integrating domain-specific VQA models with general knowledge-driven LLM agents. 
    \item \textbf{Advantage of Domain-Specific Learning for VLMs}: General VLMs that have undergone learning on domain-specific datasets, such as our aligned model and the SFT model, exhibit superior performance compared to those without such training (e.g., LLaVA, Qwen2-VL, and Deepseek-VL2). This trend highlights the limitations of general base models in specific domains and affirms the benefit of domain-adaptive training. 
    \item \textbf{Effectiveness of MoColl framework}: Models such as LLaVA-Med, LLaVA-1.5-7B, and the aligned VLM, when integrated as VQA tools within our MoColl framework and managed by an agent, demonstrate almost significantly improved performance metrics over their direct use as general VLMs. This enhancement substantiates the MoColl framework's effectiveness, illustrating its capability to optimize and extend the utility of existing models beyond their conventional applications. 
    \item \textbf{Effectiveness of Agent-Guided Tuning}: In our model collaboration framework, the aligned VLM, which has been adapted to domain-specific datasets through alignment processes, outperforms the general LLaVA models that lack domain-specific training. This validates the importance of tailoring models to specific domains. Furthermore, our tuned VQA model, which builds upon the domain-aligned VLM and is further refined through our agent-guided tuning algorithm, exceeds the performance of the aligned VLM. This affirms the effectiveness of our tuning algorithm.
\end{itemize}

These results validate the proposed MoColl framework's capacity to bridge the gap between domain-specific accuracy and general adaptability, making it a promising approach for complex image captioning tasks.

\subsection{Quantitative Results}
We show a quantitative example from IU-Xray dataset as follows. Specifically, we provide a visualized comparison of diagnostic reports generated using our MoColl framework with the agent-guided tuned VQA model and its interactive question-answer process. We juxtapose this report with one produced by the aligned VLM, comparing both to the ground truth. In the visualization, keywords accurately generated by both methods are highlighted in \colorbox{lime!50}{green}, whereas keywords uniquely and successfully generated by our method are marked in \colorbox{cyan!30}{blue}. 

\begin{table}[htp]
\begin{minipage}{0.99\textwidth}
    \centering
\scalebox{0.88}{
    \begin{tabular}{lp{10.8cm}}
    \toprule
        \multicolumn{2}{l}{\bf{Quantitative example from IU-Xray dataset}} \\ \midrule
        Ground truth & \colorbox{lime!50}{no acute cardiopulmonary abnormality}. there is a prominent calcified head to the right anterior first rib. the \colorbox{cyan!30}{aorta is tortuous}. there are \colorbox{cyan!30}{t-spine osteophytes}. the \colorbox{lime!50}{cardiomediastinal silhouette} and pulmonary vasculature are within normal limits. there is \colorbox{lime!50}{no pneumothorax or pleural effusion}. there are \colorbox{cyan!30}{no focal areas of consolidation}. \\ \midrule
        
        the Aligned VLM & \colorbox{lime!50}{no acute cardiopulmonary} process. no evidence of pneumonia. the lungs are clear. the \colorbox{lime!50}{cardiac silhouette} is normal in size. the \colorbox{lime!50}{mediastinal} and hilar contours are normal. there is \colorbox{lime!50}{no pleural effusion or pneumothorax}. \\ \midrule
        
        MoColl + tuned VQA model & \colorbox{lime!50}{No acute cardiopulmonary} process. There is mild \colorbox{cyan!30}{tortuosity} of the thoracic \colorbox{cyan!30}{aorta}. There are \colorbox{cyan!30}{T-spine osteophytes} present. The lungs are clear. There is no \colorbox{cyan!30}{consolidation} or hyperinflation. The \colorbox{lime!50}{cardiomediastinal silhouette} is normal. There is \colorbox{lime!50}{no pleural effusion or pneumothorax}.  \\ \cmidrule{2-2}
        
        \multicolumn{1}{r}{Question 1} & Can you provide details about the lung fields? \\
        \multicolumn{1}{r}{Answer 1} & no acute cardiopulmonary process. the lungs are clear. the cardiomediastinal silhouette is normal. there is no pleural effusion or pneumothorax. \\
        \multicolumn{1}{r}{Question 2} & Are there any signs of consolidation, hyperinflation, or tortuosity? \\
        \multicolumn{1}{r}{Answer 2} & the lungs are clear with no signs of consolidation or hyperinflation. there is mild tortuosity of the thoracic aorta. \\
        \multicolumn{1}{r}{Question 3} & Are there any findings related to the thoracic spine? \\
        \multicolumn{1}{r}{Answer 3} & there are T-spine osteophytes present, indicating some degenerative changes in the thoracic spine. \\
    \bottomrule
    \end{tabular}}
    \label{tab:quant}
\end{minipage}
\end{table}

This visual comparison underscores that our approach not only yields more precise but also more comprehensive diagnostic reports, demonstrating the nuanced understanding our model brings to image captioning.

\subsection{Ablation Studies}
In this subsection , we conduct comprehensive ablation studies to demonstrate the scalability and effectiveness of our proposed framework. Specifically, we study two strategies for prompt engineering in \cref{subsec:abl-ICL} and \cref{subsec:abl-length}, and two strategies for the tuning algorithm in \cref{subsec:abl-select} and \cref{subsec:abl-size}. All ablation studies are carried out within the context of cross-dataset generalization. 

\subsubsection{In-context Learning.} 
\label{subsec:abl-ICL}
\begin{figure}
    \centering
    \includegraphics[width=1.0\linewidth]{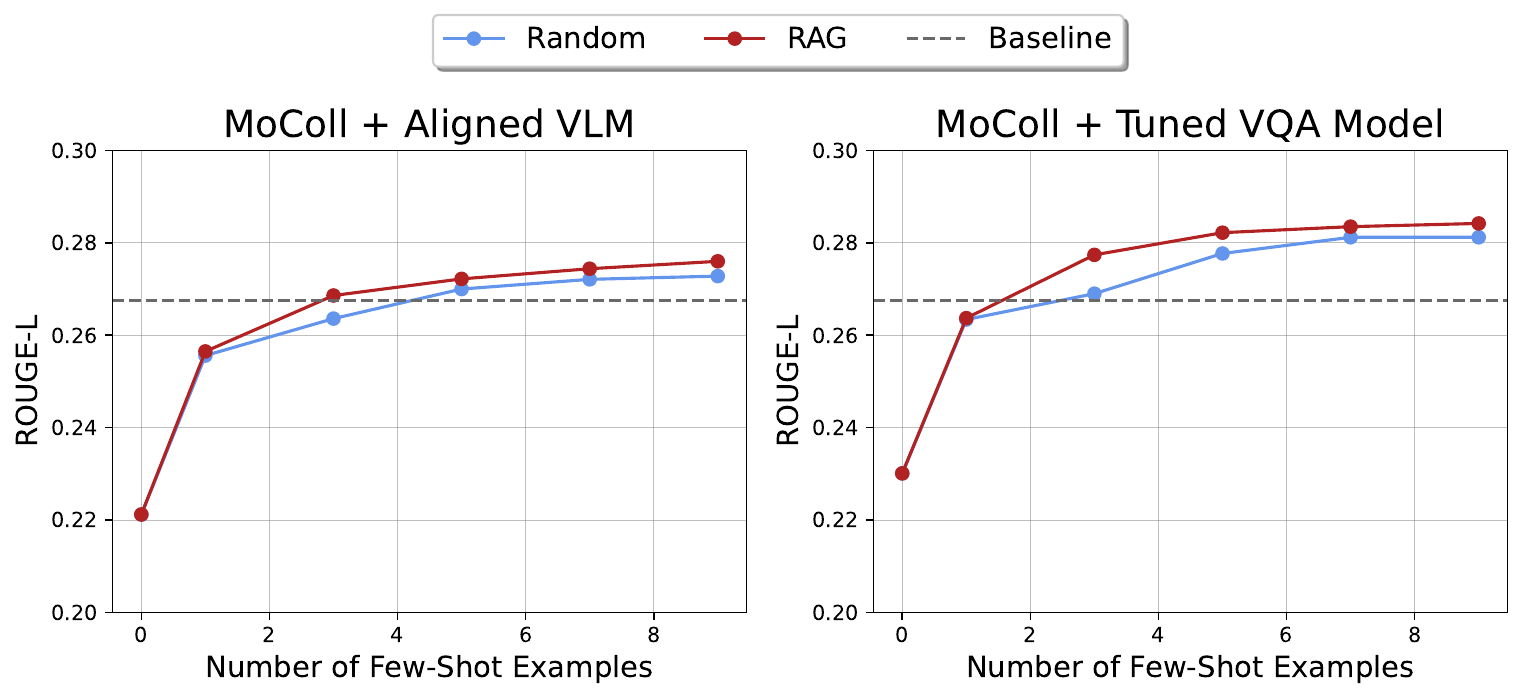}
    \vspace{-10pt}
    \caption{Ablation study on ICL. We show the ROUGE-L score with respect to the number of few-shot examples and two different example choice strategies. The left figure is for our MoColl framework with the aligned VLM, and the right one is for the MoColl framework with our agent-guided tuning algorithm. The baseline is the best score of all competing methods. }
    \label{fig:exp-ICL}
\end{figure}

In-context learning  is a pivotal technique in prompt engineering for enhancing the efficacy of model responses. In \cref{fig:exp-ICL}, we evaluate different ICL strategies and their impact on captioning performance. The results indicate that increasing the number of few-shot examples consistently improves captioning outcomes. This enhancement highlights the significant role few-shot examples play in refining the agent's querying and captioning capabilities. Furthermore, selecting examples based on image similarity, rather than randomly, tends to yield better results, since RAG strategy more likely aligns the few-shot examples with the ground truth captions of the query images. 

\subsubsection{Length of Conversation}
\label{subsec:abl-length}
\begin{figure}
    \centering
    \includegraphics[width=1.0\linewidth]{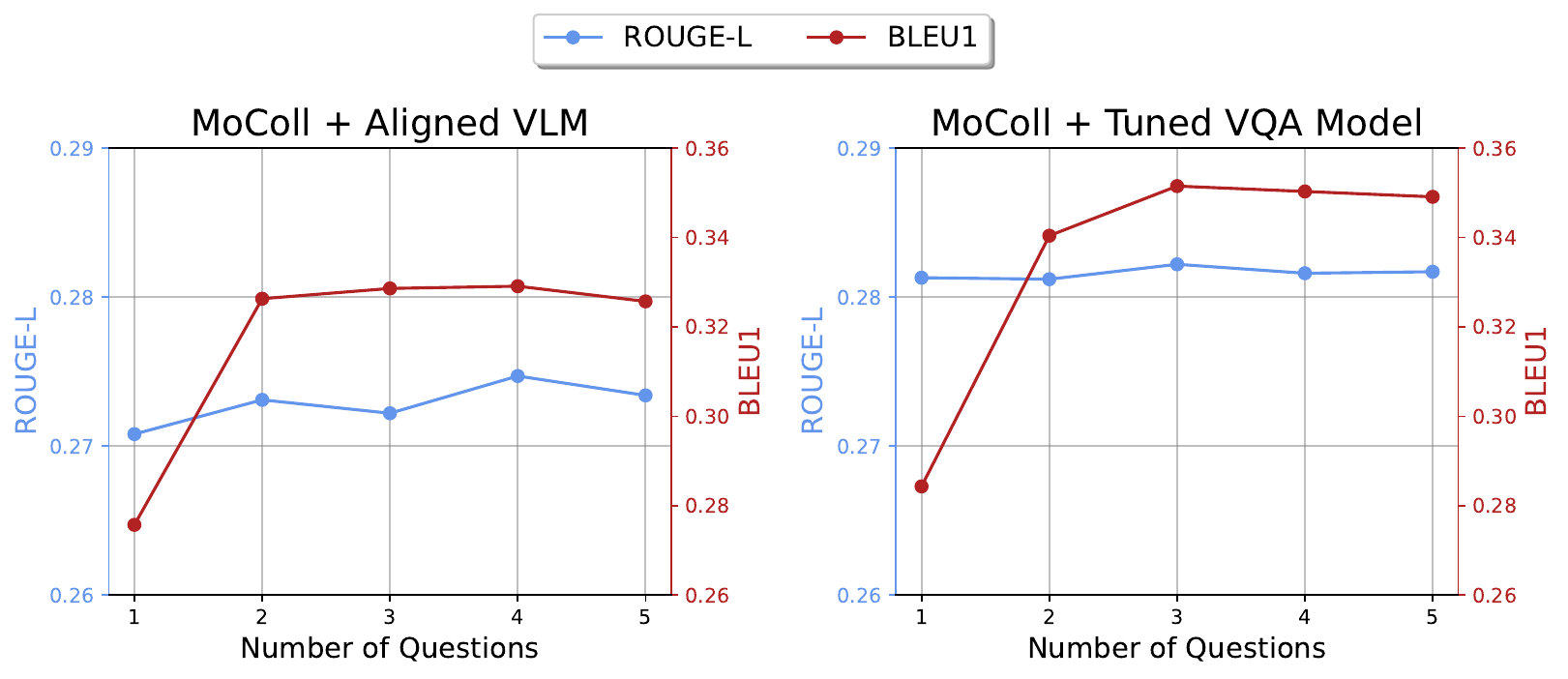}
    \vspace{-10pt}
    \caption{Ablation study on length of the chain of question-answer. We show the ROUGE-L score (blue) and BLEU1 score (red) with respect to the maximum number of questions an agent is allowed to ask. }
    \label{fig:exp-CoQA}
\end{figure}

We explore the impact of the length of conversation, i.e. the maximum number of questions an agent is allowed to ask. In \cref{fig:exp-CoQA}, we show how varying the maximum number of questions affects captioning performance. The results indicate that while extending the length of conversation has little effect on ROUGE-L scores, it significantly enhances BLEU-1 scores. This suggests that increasing the number of questions does not substantially alter the overall sentence structure of the captions but does enhance vocabulary precision. Combined with the quantitative results, our analysis further shows that the agent typically initiates the process with a broad question to grasp the global context of the images, followed by more specific queries catching details, thereby boosting the accuracy of the captions. 

\subsubsection{Selection Strategy}
\label{subsec:abl-select}
Selection is a common and effective technique to improve the quality of the data generated. In \cref{tab:exp-selection}, we show the captioning performance across various selection strategies. Our findings highlight the impact of each approach: (1) Retaining all synthetic VQA data without any selection can lead to model collapse, i.e., performing even worse than a baseline model that has not been fine-tuned at all. (2) While selecting QA pairs based on the top-performing conversations is more helpful than not selecting at all, it is not foolproof, since it may select both accurate and erroneous question-answer pairs within the same conversation with high scores. Our agent-driven approach to selectively retain correct QA pairs excels by enabling precise filtering, leading to superior model performance and reliability.

\begin{table}[t]
    \centering
    \caption{Ablation study on selection strategy. Beyond our MoColl framework, we show the ROUGE-L scores comparing three distinct strategies for selecting VQA data: (1) \textbf{No selection}: All synthetic VQA data is retained without any filtering. (2) \textbf{Top-$\bf{r}$\% ROUGE-L selection} \textit{for conversations}:Only the question-answer pairs from conversations corresponding to the top-$r$\% captions based on ROUGE-L scores are selected. (3) Our \textbf{agent-based selection} \textit{for correct question-answer pairs}: The agent selectively retains correct question-answer pairs as per its judgment. The baseline model is the aligned VLM that has not undergone fine-tuning with any VQA data. }
    \begin{tabular}{l|ccc|c}
    \toprule
        \bf{VQA data for tuning} & \bf{Is\_selection} & \bf{Criteria} & \bf{Selection\_ratio} & \bf{ROUGE-L} \\ \midrule
        No selection & \ding{55} & - & 100\% & 0.2640 \\ 
        Top-50\% ROUGE-L selection & \ding{51} & metric & 50\% & 0.2681 \\ 
        Top-25\% ROUGE-L selection & \ding{51} & metric & 25\% & 0.2701 \\ 
        Top-12.5\% ROUGE-L selection & \ding{51} & metric & 12.5\% & 0.2672 \\ 
        Agent-based selection (Ours) & \ding{51} & agent & 14.1\% & \bf{0.2822} \\ \midrule
        No VQA data (baseline) & - & - & 0\% & 0.2722 \\
    \bottomrule
    \end{tabular}
    \label{tab:exp-selection}
\end{table}

\subsubsection{Synthetic Data Size}
\label{subsec:abl-size}
\begin{figure}[t]
    \centering
    \includegraphics[width=0.5\linewidth]{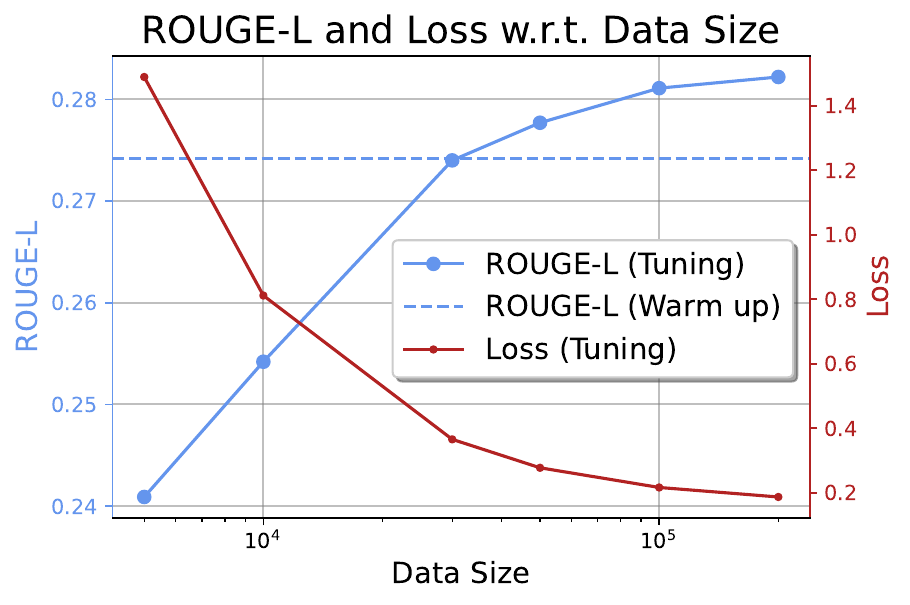}
    \vspace{-10pt}
    \caption{Ablation study on the data size. We show the ROUGE-L score (blue) and training loss (red) of our agent-guided tuning process with respect to the the size of image-caption pairs. We also show the baseline ROUGE-L score of our model collaboration framework with the aligned VLM which is trained through the warm-up stage with the whole training dataset. This baseline indicates the captioning performance at the initial fine-tuning state. }
    \label{fig:exp-size}
\end{figure}

The size of the synthetic question-answer pairs plays a critical role in the tuning of the VQA model. In \cref{fig:exp-size}, we show how the number of image-caption pairs impacts the captioning performance. Our findings indicate that as the size of the training data increases, there is a consistent decrease in loss and an increase in ROUGE-L scores. It's worth noting that, when the data size is low (less than $3 \times 10^4$), the performance of the fine-tuned VQA model falls below that of the aligned model, i.e., the initial state before fine-tuning. This suggests that a sufficient data size is essential for the success of agent-guided tuning.

\section{Conclusion}
\label{sec:conc}
In this paper, we introduced the MoColl framework, a novel approach to image captioning that utilizes a novel agent-enhanced model collaboration strategy. Our framework effectively merges domain-specific visual analysis with broad contextual knowledge by employing a VQA model as a specialized tool and an LLM as an orchestrating agent, setting it apart from traditional encoder-decoder and vision-language models. Furthermore, the agent-guided tuning algorithm stands out as a critical innovation. This algorithm enhances the VQA model by filtering synthetic data, thereby improving its adaptation for domain-specific questions from the agent. Our experimental results, particularly in the challenging domain of radiology report generation, validate the efficacy of our MoColl framework and agent-guided tuning algorithm. Our methods surpass existing models in caption accuracy, which attributes to the agent's capability to guide a structured question-answer process through the VQA model, thus refining the performance iteratively. 

Our method provides new insights for research into intelligent systems that can integrate detailed domain expertise with expansive contextual understanding. However, its effectiveness is somewhat dependent on the availability of substantial domain-specific data for fine-tuning. This requirement may pose challenges in low-resource scenarios where such data are scarce. To address these limitations, future work could explore developing more efficient data utilization techniques or incorporating few-shot learning strategies. Further, expanding the framework’s application to other multi-modal tasks could significantly broaden its utility and impact, demonstrating its adaptability across various domains.

\section*{Acknowledgment}
This work is supported in part by the New Cornerstone Investigator Program.

\bibliography{refbib.bib}
\bibliographystyle{plainnatnourl}

\newpage


\appendix

\section{Supplemental Related Work}
\label{appen-sec:relat}
\subsection{Radiology Report Generation for Chest X-ray}
\label{appen-subsec:relat-report}
Radiology report generation for chest X-rays is a specialized area of image captioning that focuses on producing detailed and clinically accurate textual reports from medical images. This task plays a crucial role in assisting radiologists, enhancing diagnostic efficiency, and improving patient care by providing consistent and comprehensive reports.

With the advancement of deep learning, encoder-decoder architectures became prevalent. Models like \cite{Shin_2016_CVPR} and \cite{Wang_2017_CVPR} utilized Convolutional Neural Networks (CNNs) for encoding image features and Recurrent Neural Networks (RNNs) for decoding them into text. Attention mechanisms were later integrated to allow models to focus on specific regions of the image, as seen in works by \cite{chen-emnlp-2020-r2gen} and \cite{chen-etal-2021-cross-modal}, improving the relevance and accuracy of the generated reports. Beyond these standard architectures, more advanced models have been proposed specifically for radiology tasks such as \cite{yang2021joint} and \cite{wang2022cross}. 

To incorporate medical knowledge, some studies integrated clinical knowledge bases into their models. For example, \cite{Li_Liang_Hu_Xing_2019} introduced a model that aligns image features with medical concepts, ensuring the use of correct terminology and facilitating the generation of more informative reports. Additionally, \cite{yang2023radiology} adopted a multi-task knowledge transfer mechanism to enhance the generation process by simultaneously learning disease classification and report generation tasks.

Despite these advancements, existing models often struggle to fully leverage both the general knowledge of large-scale models and the domain-specific expertise required for medical reporting. Challenges such as generating coherent narratives, accurately reflecting clinical findings, and adapting to varied patient cases remain. Our method is capable of addressing these challenges by combining the expansive knowledge and language generation abilities of LLMs with the specialized visual processing capabilities of a domain-specific VQA model. Therefore, we use the radiology report generation task as an example to run experiments, demonstrating the effectiveness of our approach.

\section{Implementation Details}
\label{appen-sec:imp}
\subsection{Detailed Setup}
\label{appen-subsec:setup}
Our proposed approach is implemented using Pytorch. The main parameters of the server are listed below: the operating system is Rocky Linux 8.8, the CPU is Intel Xeon Platinum 8358P with 2.60 GHz and 64 cores, the GPU is an 8-card A800 80G, and the memory capacity is 512 GB. 

\subsection{Data Prepossessing}
\label{appen-subsec:data}
For both datasets, we use `finding + impression' as the radiology report for each report. We split the IU-Xray into train and test set by 8:2 randomly and MIMIC-CXR according to its official split. We filter out any missing data, such as reports without corresponding images, or images without corresponding reports. Each report corresponds to a minimum of one image and a maximum of four images. Overall, there are 767 test data in IU-Xray, and 222758 training data and 3269 test data in MIMIC-CXR. To align with the input shape of the CLIP Vision Encoder, we clip all images to 224*224. Following the \cite{chen-emnlp-2020-r2gen}'s official implementation of report preprocessing, the cut-off frequency for the words is set as 3 for IU-Xray and 10 for MIMIC-CXR. 

\subsection{Our Methods}
\label{appen-subsec:method}

\paragraph{MoColl framework.} For the MoColl framework, we set the maximum number of questions an agent is allowed to ask as 6 in the within-dataset evaluation and 3 in the cross-dataset generalization. In all experiments, except those specifically stated, we always use the RAG as the ICL strategy for the agent. We always provide 5 few-shot examples of captions for the agent. Notice that we can still use few-shot examples with the MIMIC-CXR training dataset in the setup of cross-dataset generation. we always provide 5 few-shot examples of captions to enhance the understanding of the specific domain. For the generation setting, we set the max length of tokens as 4096 and temperature as 0 for both VQA model and LLM-based agent. 

\paragraph{Warm-up stage.} In the warm-up stage, we keep both the visual encoder and LM weights frozen, and only update the projection network. We only train 1 epoch with auto-regressive loss. We set the learning rate as 2e-3 and use cosine learning rate scheduler with warm-up ratio as 0.03. The max length of tokens is 4096, and the batch size is 16 per device. 

\paragraph{Agent-guided tuning stage.} In the agent-guided tuning stage, we show the implementation details of the inference step, selection step, and training step individually as follows. For the inference step, we also set the max length of tokens as 4096 and temperature as 0.1 for both VQA model and LLM-based agent. The few-shot examples are captions randomly selected from the training set. For the selection step, we set the temperature of the selection agent as 0. We can also improve the questioning prompt according to the conversation manually. For the training step, the visual encoder weights are still frozen, and the LM and the projection network can be updated. We train 5 epochs in the within-dataset evaluation and only 1 epoch in the cross-dataset generalization to prevent overfitting. The loss function is auto-regressive loss with weighted KL penalty to enhance stability. We set the learning rate as 3e-7 and use cosine learning rate scheduler with warm-up ratio as 0.03. The max length of tokens is 4096, the batch size is 4 per device and the gradient accumulation steps are 2. 

Other hyperparameters remain their default values. 

\subsection{Competing Methods}
\label{appen-sec:comp}

\subsubsection{Specialized models}
For all specialized models, we leverage their official code repository to reproduce the results based on our setup. Especially, for XProNet \cite{wang2022cross} and M2KT \cite{yang2023radiology}, we follow their preprocessing process and use CheXbert \cite{smit2020chexbert} to rebuild labels with reports under our setup. We set the maximum sequence length as 530, which is the maximum length of captions in our experiments. Other hyperparameters remain default values. 

\subsubsection{General VLMs}
We use official implementations and released model checkpoints for LLaVA-Med and LLaVA-v1.5-7B, and official APIs for Qwen2-VL-7B-Instruction and Deepseek-VL2. The aligned VLM is trained through the warm-up stage. The fine-tuned VLM is trained on the image-instruction-caption tuple with auto-regressive loss. Its visual encoder weights are still frozen, and the LM and the projection network can be updated. We set its learning rate as 3e-7 and use cosine learning rate scheduler with warm-up ratio as 0.03. 

For the generation setting, we provide 5 few-shot examples of captions randomly to enhance the understanding of the specific domain. We also set the temperature as 0 and other hyperparameters remain the default values. 

\subsubsection{MoColl framework with VQA models}
We also use official implementations and released model checkpoints for LLaVA-Med and LLaVA-v1.5-7B, serving as VQA tool models instead of general VLMs working independently. 


\end{document}